\documentclass[sigconf,authordraft,review=false,timestamp=false,nonacm=true]{acmart}

\setcopyright{acmlicensed}
\copyrightyear{2024}
\acmYear{2024}
\acmDOI{XXXXXXX.XXXXXXX}

\acmConference[PASC '24]{Platform for Advanced Scientific
Computing Conference }
{June 03--05,2024}
{Zurich, Switzerland}

\input{packages}

\newcommand{\diff}[1]{\ensuremath{\operatorname{d}\!{#1}}}

\def\0{\bm{0}}
\def\1{\bm{1}}
\newcommand{\train}{\mathcal{S}}

\def\rmF{{\mathbf{F}}}

\def\rmS{{\mathbf{S}}}

\def\rmW{{\mathbf{W}}}

\def\vtheta{{\bm{\theta}}}

\def\vg{{\bm{g}}}

\def\vp{{\bm{p}}}

\def\vw{{\bm{w}}}
\def\vx{{\bm{x}}}
\def\vy{{\bm{y}}}
\def\vz{{\bm{z}}}

\newcommand{\E}{\mathbb{E}}

\newcommand{\R}{\mathbb{R}}

\DeclareMathOperator*{\argmin}{arg\,min}

\newcommand{\params}{\ensuremath{\bm{\vartheta}}} 
\newcommand{\net}{\ensuremath{f_{\params}}} 

\newcommand{\grad}{\nabla}

\newcommand{\loss}{\ensuremath{\mathcal{L}}}

\newcommand{\payoff}{\ensuremath{\upnu}}
\newcommand{\callprice}{\ensuremath{V_C}}

\newcommand{\normalpdf}{\ensuremath{\varphi}}
\newcommand{\normalcdf}{\ensuremath{\Phi}}
\newcommand{\inparamsset}{\ensuremath{\Theta_{\text{in}}}}
\newcommand{\outparamsset}{\ensuremath{\Theta_{\text{out}}}}

\begin{document}


\title{Towards Sobolev Pruning}

\author{Neil Kichler}
\authornote{Both authors contributed equally to this research.}
\email{kichler@stce.rwth-aachen.de}
\affiliation{%
  \institution{RWTH Aachen University}
  \city{Aachen}
  \country{Germany}
}

\author{Sher Afghan}
\authornotemark[1]
\email{afghan@stce.rwth-aachen.de}
\affiliation{%
  \institution{RWTH Aachen University}
  \city{Aachen}
  \country{Germany}
}

\author{Uwe Naumann}
\email{naumann@stce.rwth-aachen.de}
\affiliation{%
  \institution{RWTH Aachen University}
  \city{Aachen}
  \country{Germany}
}

\begin{abstract}
The increasing use of stochastic models for describing complex phenomena warrants surrogate models that capture the reference model characteristics at a fraction of the computational cost, foregoing potentially expensive Monte Carlo simulation. The predominant approach of fitting a large neural network and then pruning it to a reduced size has commonly neglected shortcomings. 
The produced surrogate models often will not capture the sensitivities and uncertainties inherent in the original model. In particular, (higher-order) derivative information of such surrogates could differ drastically.  
Given a large enough network, we expect this derivative information to match. However, the pruned model will almost certainly not share this behavior.

In this paper, we propose to find surrogate models by using sensitivity information throughout the learning and pruning process. We build on work using Interval Adjoint Significance Analysis for pruning and combine it with the recent advancements in Sobolev Training to accurately model the original sensitivity information in the pruned neural network based surrogate model. We experimentally underpin the method on an example of pricing a multidimensional Basket option modelled through a stochastic differential equation with Brownian motion. The proposed method is, however, not limited to the domain of quantitative finance, which was chosen as a case study for intuitive interpretations of the sensitivities.
It serves as a foundation for building further surrogate modelling techniques considering sensitivity information.
\end{abstract}



\begin{teaserfigure}
  \centering
  \includegraphics[width=\textwidth]{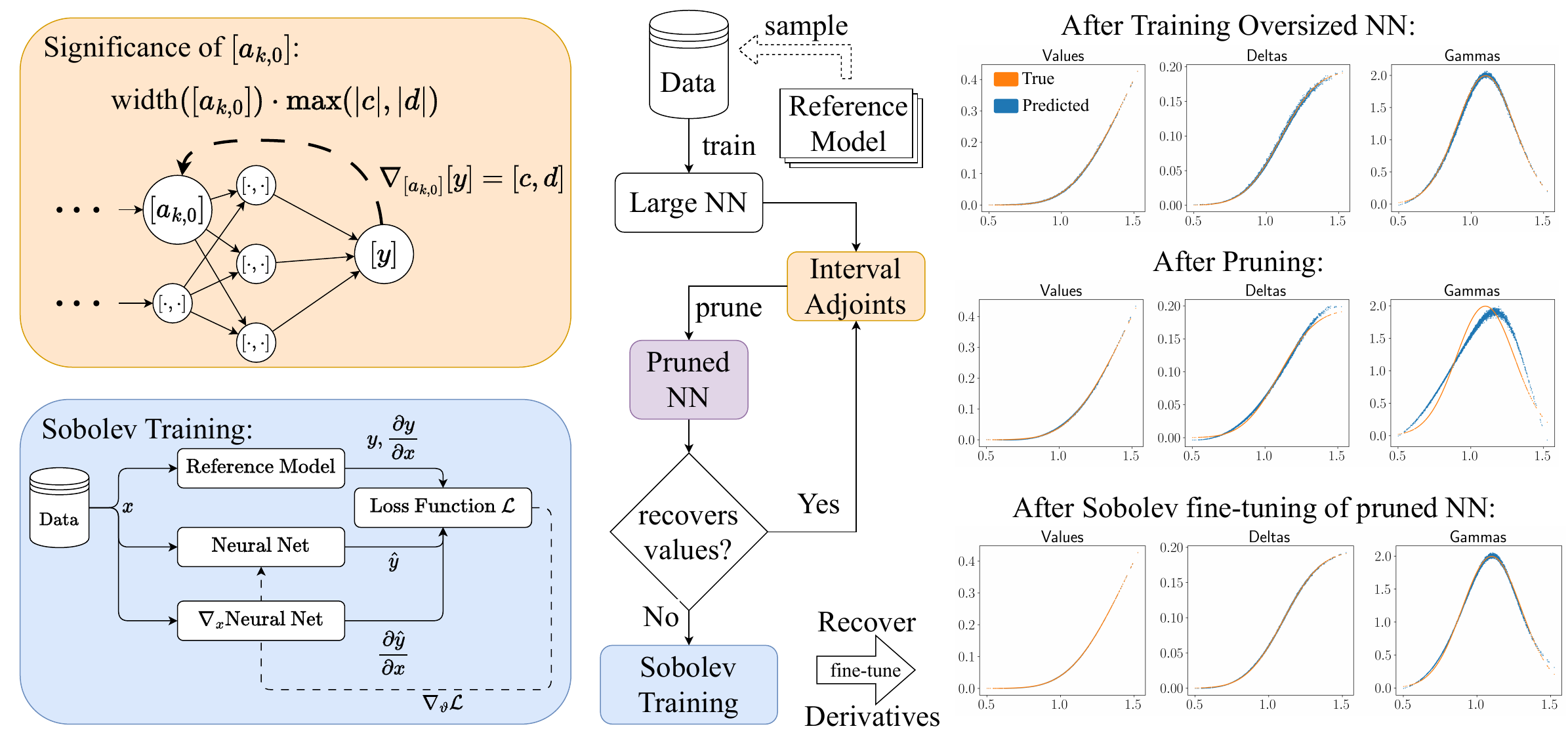}
  \caption{Train a large neural network, prune it efficiently using Interval Adjoint Significance Analysis to smallest size recovering predicted values, and apply Sobolev Training for final fine-tuning, recovering derivative information.}
  \Description{A diagram of the proposed method. In the center, a flow chart highlights the steps of the procedure. Beginning on the top, we train a neural network with data generated from a reference model. The next block is interval adjoints, indicating the use of it for model pruning which is executed iteratively until the pruned model cannot recover the targets to predict. The final block is Differential ML to denote its use for the final fine-tuning and recovering of the sensitivity information. On the left are further details of Interval Adjoints and Differential ML displayed. On the right are the prediction results shown for the oversize neural network, the pruned one (which is worse), and the pruned one after fine-tuning with Differential ML (similar to original network).}
  \label{fig:teaser}
\end{teaserfigure}

\received{1 December 2023}

\maketitle

\section{Introduction}
High-dimensional stochastic models are necessary for modeling complex phenomena of the real world. 
Fields like biology, material science, and quantitative finance, among many others, embrace using a stochastic process during simulation, as it is most often infeasible to rely on the fundamental governing physical equations. The techniques of stochastic modelling that were initially developed in the field of statistical mechanics led to many domains that can now reason about the macro perspective without getting stuck in an intractable simulation of a more fundamental process.

One large general class of stochastic numerical methods using Monte Carlo (MC) sampling are, however, particularly computationally expensive in converging to the conditional expectation \cite{glasserman2003}.
Nonetheless, MC often remains the only applicable option.
Consider, as an example, option pricing models in quantitative finance.
For simple models and simple payoff functions, analytic solutions may exist. But, in general, using more interesting models or exotic payoff functions will lead to models without analytic solutions. It remains to fall back to expensive MC simulation. 
If a model is not fast enough, we aim to find surrogate models that execute much faster while staying sufficiently accurate.

The advancements in Machine Learning (ML) provide a promising avenue for the creation of much faster, yet accurate surrogate models. Especially in high-dimensional settings, neural network based methods excel because, in many occasions, they can overcome the curse of dimensionality. 
Such surrogate models can be used in production settings where there exists a need for near real time analysis, prototyping, or simulation. However, they are prone to overfitting and usually require some form of regularization, which will introduce a potential bias. Moreover, they require large training data sets and perform poorly in recovering the underlying uncertainties \cite{czarnecki2017sobolev, Huge2020}. 

By adopting a stochastic perspective, one, however, introduces various uncertainties into the computation and simulation of the process.
Therefore, quantifying uncertainty is quickly becoming an indispensable part of all scientific fields. 
The field of uncertainty quantification is vast 
and far-reaching, yet is often not fully considered in neural network pruning and surrogate modeling. 
Prominent techniques include the use of derivative information
or Interval Arithmetic \cite{ref_interval_math_2, IASAFF}.
Combining those methods has, to the best of our knowledge, not been considered so far and could open interesting avenues.

Overall, what is currently missing is to learn from differential data throughout the entire process of training and pruning surrogates. In particular, samples of the sensitivity at certain critical points can ensure that the surrogate models follow the uncertainties of the reference problem \cite{czarnecki2017sobolev, Huge2020}. In this paper, we systematically explore the landscape of incorporating sensitivity information into the process of pruning existing and finding new surrogate models.
In the context of training, the technique of incorporating derivative information into the loss is already known by Sobolev Training \cite{czarnecki2017sobolev} and is loosely based on the concept of the Sobolev space. For pruning, we now aim to follow suite and want to move towards Sobolev pruning.

\subsection{Finding surrogate models}

When trying to find a small surrogate model, one can approach the problem in two broad ways. 

\paragraph{From large to small surrogates}
We can start training a sufficiently large neural network found through experimentation, experience and random guessing of the model size and architecture to then prune and quantize it back to a smaller surrogate model.

\paragraph{Remain small throughout learning}

Alternatively, one can start as small as possible and train a network, only growing larger when the accuracy remains insufficient.
Knowing how small a network could be is, however, a challenge in itself and often leads to random experimentation, nonetheless.
\newline\newline\noindent
Since the former is usually more common - as it is more likely to succeed - we start in the same manner. However, we instead want to use the pruned network as information of an appropriate neural network model size. Then further fine-tuning will be achieved through a computationally more expensive learning process that incorporates (second-order) differential data. During this training, the pruned model can recover the proper uncertainty information and perform beyond surrogate models obtained through typical pruning methods on neural networks.

\subsection{Main contributions}
The main contributions of this paper are as follows:

\begin{itemize}
    \item We compute a surrogate model by pruning a larger neural network using Interval Adjoint Significance Analysis.
    \item We, furthermore, use Sobolev Training for fine-tuning the pruned surrogate model and hence recover first-order and second-order derivative prediction accuracy.
    \item Although the individual techniques are not new, their combination and the overall plan we are laying out is, to the best of our knowledge, novel.
    \item We provide an experimental comparison of the presented methods on a Gaussian basket option pricing model.
\end{itemize}

The approach is broadly visualized in \autoref{fig:teaser}. It requires various techniques that get incrementally introduced.
In \autoref{sec:diffml}, we introduce Sobolev Training as a way to learn from derivative information.
We then consider pruning using Interval Adjoint Significance Analysis in \autoref{sec:pruning}. \autoref{sec:case-study} introduces the case study of a Gaussian basket pricing model and \autoref{sec:least-squares-mc} explains how we can find pathwise derivatives of MC samples before we present the results of the various configurations in \autoref{sec:results}.

\section{Learning a Surrogate through derivative information}\label{sec:diffml}

Traditional neural network based learning of surrogates considers a dataset of values generated from the function representing the model we wish to approximate. However, the reference model often encodes more information than just output values given some input values. In particular, derivative information can, in many cases, easily be obtained through Algorithmic Differentiation (AD) \cite{AD2}, but is often neglected. 

One method that does not neglect derivatives is known as Sobolev Training \cite{czarnecki2017sobolev,Huge2020}.
The core idea boils down to using derivative information during training of a neural network. 
An additional loss term is considered that penalizes deviating gradient information of the neural network, resulting in the Sobolev loss.

\begin{definition}[Sobolev Loss]\label{def:differentialloss}
Given input $\vx$, target $\vy$, predicted output $\net(\vx)$, differential target $\nabla_{\vx}\vy$, and predicted differential $\nabla_{\vx}\net(\vx)$, the differential loss is defined by:
\begin{equation}\label{eq:differentialloss}
    {\lVert \vy - \net(\vx) \rVert}_{2}^{2} + \lambda {\lVert \nabla_{\vx} \vy - \nabla_{\vx} \net(\vx) \rVert}_{2}^{2},
\end{equation}
where $\lambda \in \R_{\geq0}$ is an added balancing factor.
\end{definition}

\citet{pmlr-v80-srinivas18a} highlight that the Sobolev loss arises naturally as a consequence of considering the expectation over the mean squared error (MSE) of the inputs given Gaussian noise perturbations.

\begin{figure}[h]
    \centering
    \includegraphics[width=\linewidth]{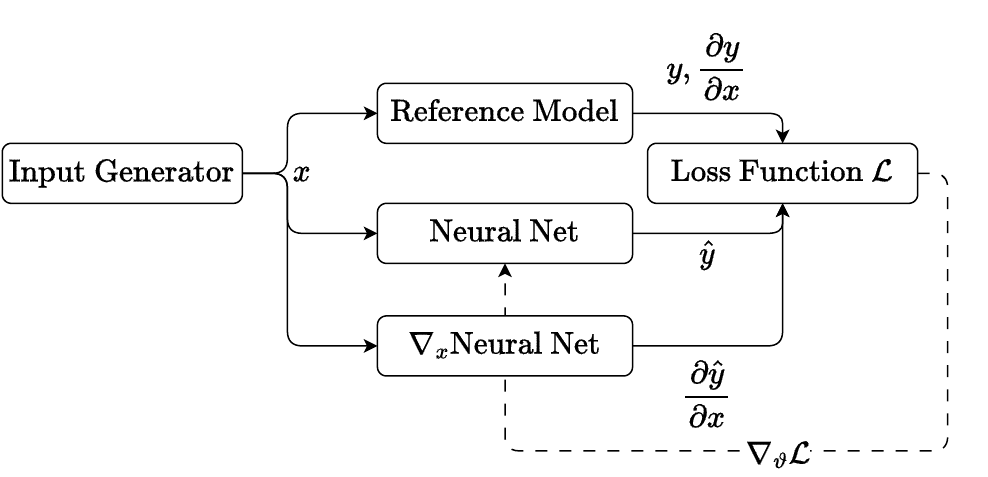}
    \caption{Visualization of Sobolev Training.}
    \label{fig:diffml-training}
\end{figure}

The remaining aspects of the training procedure follows standard ML practices and is summarized in \autoref{alg:diffml}.
Given a neural network surrogate model $\mathcal{N}(\params)$ with parameters $\params$
and learned function $\net$, a training iteration is performed as follows:
First, the reference model sampler $\train$ is used to generate training data (either during the iteration or precomputed before training). Second, the network is used to predict outputs with $\net(\vx_i)$. Using adjoint AD, we furthermore compute the gradient information with respect to the input sample $\vx_{i}$, i.e. $\nabla_{\vx}\net(\vx_i)$. Using a loss function, e.g., $\loss = {\lVert \cdot \rVert}_{2}^{2}$, the Sobolev loss is computed and used to find the mini-batch gradient  $\hat{\vg}$. A stochastic gradient based optimizer $G$, like SGD or Adam \cite{KingmaB14}, finally computes parameter updates for the surrogate model until they converge.

\begin{algorithm}
\caption{Sobolev Training \cite{czarnecki2017sobolev}.}\label{alg:diffml}
\begin{algorithmic}
\Require The following inputs must all be initialized. 
\begin{itemize}[label=\ding{251}]
    \item Surrogate model $\mathcal{N}(\params)$ with function $\net$ and parameters $\params$
    \item Reference model $\train$
    \item Optimizer $G$
\end{itemize}
\While{$\params$ not converged}
    \State $\{(\vx_i,\vy_i,\nabla_{\vx}\vy_i)\}_{i=1}^m \sim \train$ \Comment{Sample training data}
    \State $\hat{\vg} \gets \frac{1}{m} \grad_{\params} \sum_{i=1}^{m} \loss(\net(\vx_{i}), \vy_{i}) + \lambda \loss(\nabla_{\vx}\net(\vx_i),\nabla_{\vx}\vy_i)$ 
    \State $\params \gets G(\params, \hat{\vg})$ \Comment{Update parameters}
\EndWhile\\
\Return $\mathcal{N}$
\end{algorithmic}
\end{algorithm}

The proposed method can be extended to second-order differential data. However, the computation of the full Hessian is often infeasible for larger reference models. Thus, fewer (hopefully informative) directions have to be sampled instead. Alternatively, second-order information could be used in only every $k^{\text{th}}$ iteration.
A naive approach could consider random directions for the hessian vector products \cite{martens2012estimatinghessian}.
Instead, \citet{kichler2023sodml} uses information from a principal component analysis on the computed gradient data to find the principal components corresponding to the directions of maximal variance. Then, the principal components can either be used as an improved orthonormal basis compared to the Cartesian basis vectors or reduced down to the $k$ most important directions, describing, e.g., 95\% of the observed variance.

A discussion of this method is beyond the scope of this paper, but will serve useful during training and evaluation as it promises even better results, particularly in prediction of (second-order) differential data.

However, in Sobolev Training it remains unclear how large the surrogate model should be for appropriate learning to occur. In other words, how small can we make the surrogate model while retaining a given accuracy?
In practice, one often uses past experience, or randomly guesses and adjusts the model size.
Often, however, a large neural network model already exists. An alternative approach could then be to prune the model to a smaller size and retrain or fine-tune the pruned model using Sobolev Training. We therefore call it Sobolev fine-tuning. 
A particularly effective pruning method in this context is considered next.

\section{Neural Network Pruning}\label{sec:pruning}
An important factor in designing efficient neural network (NN) architectures is the selection of hyperparameters, such as the learning rate, regularization, and optimization method \cite{hyper-parameters}. One particularly important parameter is the network size. If the size is too small, the network may struggle to learn the underlying problem, while if it is too large, the network may struggle with generalization, as a result, the model may not generalize well to new, unseen data because it has essentially memorized the training set rather than learning the underlying patterns. It can be difficult to predict the optimal size of NNs beforehand.

Different methods for pruning have been developed, which can be broadly categorized into two main categories: unstructured pruning and structured pruning. 
	
Unstructured pruning ~\cite{lee2018snip, mocanu2018scalable, pruning2,pruning_grow,pruning_energy} is typically less aggressive and removes individual weights or connections rather than entire neurons or layers. It results in sparse matrices by pruning arbitrary weight connections depending upon their sensitivity to the network. Sparse algorithms are used, and indices are stored for computation. However, current GPUs and multi-core CPUs do not perform efficiently over sparse networks. There is a throughput problem and hence no efficiency gains ~\cite{pruning-str1,pruning-str2}. Typically, in dense layers, the magnitude or the $L^1$/$L^2$ norm of the weights $\vw = [w_1, w_2, w_3, ..., w_n]$ in a single layer, computed as $L^p = ||w_i||_{p}$ where typically $p=1,2$, is used for identification of less important weights in the layer. Less important weights are then removed on the basis of the ranking of their magnitude. The norm of the weight gradients could also be used to make the importance ranking of weights.  

Structured pruning ~\cite{pruning-str1,pruning-str2,pruning-channel}, on the other hand, involves removing entire neurons, channels, or filters from the network. This approach is typically more aggressive and can lead to larger reductions in the number of parameters in the model. This approach not only gives us a pruned network but also preserves the regularity in the network. Indices do not need to be remembered, and the matrices are dense.

There are various methods for determining the importance of neurons, connections, or filters in a neural network ~\cite{pruning-str1,pruning-str2,pruning-channel}. One common approach is to rank neurons based on their saliency, which is typically calculated using the sum of incoming or outgoing weights to a neuron. Another approach is to use the norm of the neuron weights as a measure of importance. Optimal brain damage (OBD) ~\cite{OBD} and optimal brain surgeon (OBS) ~\cite{OBS} are examples of pruning methods that remove neurons or connections that have the smallest impact on the overall performance of the network. These methods use second derivative information for examining how the rate of change of certain values in the network impacts its performance. However, this detailed analysis can be computationally demanding, especially when dealing with a large number of parameters, making it a bit challenging for practical use in some situations.

In this paper, we use interval arithmetic and algorithmic differentiation in order to inform the compression of a NN, as introduced in ~\cite{IASAFF}. In the following section, a concise overview of IA, AD and their combination for interval data is presented, accompanied by an example.

\subsection{Interval Arithmetic and Interval Adjoints}
	
	Interval Arithmetic (IA) ~\cite{ref_interval_math} is a branch of mathematics that deals with ranges or intervals of numbers, rather than specific numbers. It can be useful in situations where there is uncertainty or variability in the inputs or parameters of a mathematical problem. By representing numbers as intervals, rather than single values, it can provide more robust and reliable results in the face of uncertainty. IA is well defined for operations such as addition, subtraction, multiplication, and division of intervals, as well as more advanced operations. These operations and their applications are discussed in detail in ~\cite{ref_interval_math,ref_interval_math_2}. 
 
    We denote an interval through $[x]=[x_\bot, x_\top]$, with $x_\bot$, $x_\top$ the lower bound and upper bound of the interval, respectively. We denote a vector of intervals by $[\vx] = ([x]_1, [x]_2, \ldots, [x]_n)^{\top} \in [\R]^n$, and interval domain
    $[\R] := \bigl\{ [x_1, x_2] \mid x_1 \leq x_2, \text{and } x_1, x_2 \in \R \cup \{-\infty, \infty\} \bigr\}$.
    A function $f: [\R]^n \rightarrow [\R]$ can be implemented using IA. It ensures that, if all operations in $f$ are well defined in IA, $f$ covers the entire output domain, representing a superset $f([\vx]) \supseteq  \bigl\{f([\vp]) \mid [\vp] \subseteq [\vx]\bigr\}$, for the vector interval  $[\vx]$ covering the entire input domain. Furthermore, IA guarantees enclosing the output domain for all the intermediate interval variables, too.
	
	AD ~\cite{AD1,AD2} uses the chain rule of differentiation to recursively compute the derivatives of a function, by decomposing it into a sequence of elementary operations. The adjoint AD (a.k.a reverse mode) computes the function output in a forward pass, i.e. the primal section, and its derivatives with respect to each input variable in a reverse pass, i.e. the adjoint section, by recursively applying the chain rule from the output to the inputs. The adjoint mode of AD is particularly useful for computing gradients of functions with many inputs and few outputs, such as in deep learning, where the function can be a neural network with many parameters.
	Combining IA with AD, allows the computation of rigorous bounds on the range and derivatives of functions with interval inputs ~\cite{riehme2015significance}.

    If we apply adjoint AD to intervals, it not only gives us guaranteed enclosures of its intermediate and output variables but it also computes their interval partial derivatives.
\begin{figure}[h]
		\centering
		\includegraphics[width=\linewidth]{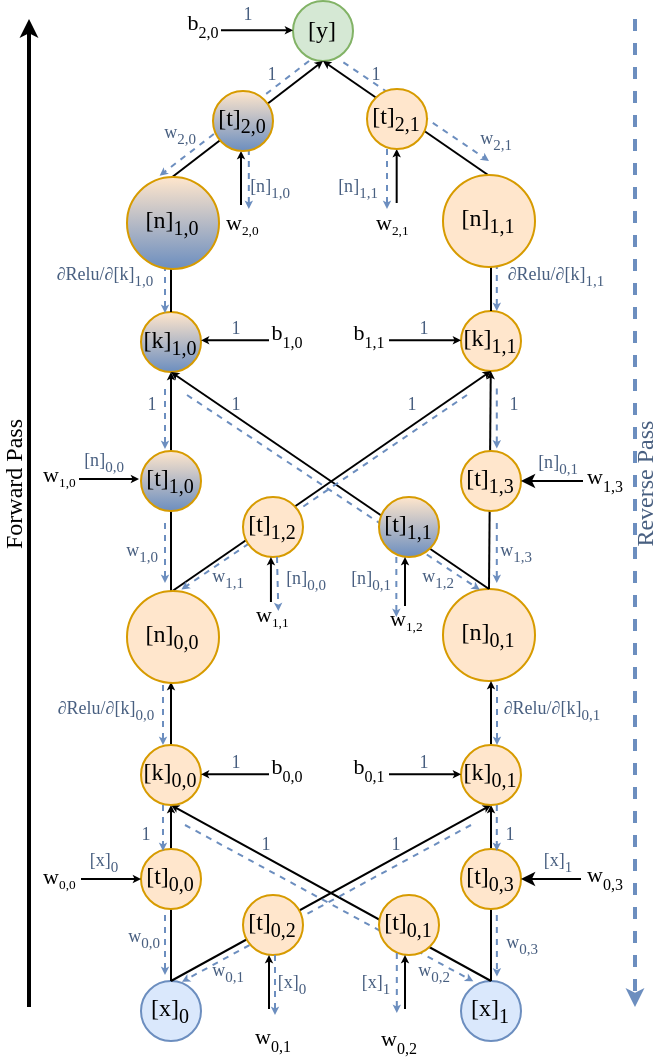}
		\caption{Computational graphs of two layer NN to evaluate primal values and interval adjoints. The inputs, intermediate variables, hidden node values and output are shown inside the circles along with the mathematical operator or the activation funtion to be applied on incoming variables. Forward pass is shown with black arrows while blue arrows and the blue values show the reverse pass and the partial derivatives.}
		\label{fig_0}%
\end{figure}

\begin{table}[h]
    \centering
    \begin{adjustbox}{width=\linewidth,center}
    \renewcommand{\arraystretch}{0.8}
    \addtolength\tabcolsep{0pt}
    \small
    \setlength{\tabcolsep}{0pt} 
    \begin{tabular}{rrllrl}
       \multicolumn{3}{l}{\small{\textbf{ Primal Trace}}} &
       \multicolumn{3}{l}{\small{\textbf{ Interval Adjoints Trace}}} \\
       \multirow{8}{*}{\begin{tikzpicture}
           \draw[thick,<-] (-0.0,0) -- (-0.0,6.5) node[anchor=south west] {};
       \end{tikzpicture}} & & &
       \multirow{8}{*}{\begin{tikzpicture}
           \draw[thick,->] (0,0) -- (0,6.5) node[anchor=south west] {};
       \end{tikzpicture}} & \\
       &$[t]_{0,0}$&= $[x]_0 \cdot w_{0,0} = [0.12544, 1.2544]$ &&$[\overline{t}]_{0,0}$&= $ [\overline{k}]_{0,0} = [0.7421, 0.7421]$\\
       &$[t]_{0,1}$&= $[x]_1 \cdot w_{0,2} = [0.0277, 0.277]$ &&$[\overline{t}]_{0,1}$&= $ [\overline{k}]_{0,0} = [0.7421, 0.7421]$\\
       &$[t]_{0,2}$&= $[x]_0 \cdot w_{0,1} = [0.1121, 1.1213]$ &&$[\overline{t}]_{0,2}$&= $ [\overline{k}]_{0,1} = [1.2733, 1.2733]$\\
       &$[t]_{0,3}$&= $[x]_1 \cdot w_{0,3} = [-0.023, -0.002]$ &&$[\overline{t}]_{0,3}$&= $ [\overline{k}]_{0,1} = [1.2733, 1.2733]$\\
       &$[k]_{0,0}$&= $[t]_{0,0} + [t]_{0,1} + b_{0,0} = [0.315, 1.693]$ &&$[\overline{k}]_{0,0}$&= $ \partial \text{ReLU}/\partial [k]_{0,0} \cdot [\overline{n}]_{0,0}$\\
       &$[k]_{0,1}$&= $[t]_{0,2} + [t]_{0,3} + b_{0,1} = [0.368, 1.397]$ &&$[\overline{k}]_{0,1}$&= $ \partial \text{ReLU}/\partial [k]_{0,1} \cdot [\overline{n}]_{0,1}$\\
       &$[n]_{0,0}$&= $\text{ReLU}([k]_{0,0}) = [0.3151, 1.6934]$ &&$[\overline{n}]_{0,0}$&= $w_{1,1} \cdot [\overline{t}]_{1,2} + w_{1,0} \cdot [\overline{t}]_{1,0}$\\
       &$[n]_{0,1}$&= $\text{ReLU}([k]_{0,1}) = [0.3677, 1.3974]$ &&$[\overline{n}]_{0,1}$&= $w_{1,3} \cdot [\overline{t}]_{1,3} + w_{1,2} \cdot [\overline{t}]_{1,1}$\\
       &$[t]_{1,0}$&= $[n]_{0,0} \cdot w_{1,0} = [0.0496, 0.2665]$ &&$[\overline{t}]_{1,0}$&= $ [\overline{k}]_{1,0} = [0.4287, 0.4287]$\\
       &$[t]_{1,1}$&= $[n]_{0,1} \cdot w_{1,2} = [0.098, 0.3725]$ &&$[\overline{t}]_{1,1}$&= $ [\overline{k}]_{1,0} = [0.4287, 0.4287]$\\
       &$[t]_{1,2}$&= $[n]_{0,0} \cdot w_{1,1} = [0.1556, 0.8361]$ &&$[\overline{t}]_{1,2}$&= $ [\overline{k}]_{1,1} = [1.3663, 1.3663]$\\
       &$[t]_{1,3}$&= $[n]_{0,1} \cdot w_{1,3} = [0.3119, 1.1854]$ &&$[\overline{t}]_{1,3}$&= $ [\overline{k}]_{1,1} = [1.3663, 1.3663]$\\
       &$[k]_{1,0}$&= $[t]_{1,0} + [t]_{1,1} + b_{1,0} = [0.241, 0.733]$ &&$[\overline{k}]_{1,0}$&= $ \partial \text{ReLU}/\partial [k]_{1,0} \cdot w_{2,0}$\\
       &$[k]_{1,1}$&= $[t]_{1,2} + [t]_{1,3} + b_{1,1} = [0.766, 2.321]$ &&$[\overline{k}]_{1,1}$&= $ \partial \text{ReLU}/\partial [k]_{1,1} \cdot w_{2,1}$\\
       &$[n]_{1,0}$&= $\text{ReLU}([k]_{1,0}) = [0.241, 0.733]$ &&$[\overline{n}]_{1,0}$&= $w_{2,0} = [0.4287, 0.4287]$\\
       &$[n]_{1,1}$&= $\text{ReLU}([k]_{1,1}) = [0.7664, 2.3205]$ &&$[\overline{n}]_{1,1}$&= $w_{2,1} = [1.3663, 1.3663]$\\
       &$[t]_{2,0}$&= $[n]_{1,0} \cdot w_{2,0} = [0.1034, 0.3141]$ &&$[\overline{t}]_{2,0}$&= $[1, 1]$\\
       &$[t]_{2,1}$&= $[n]_{1,1} \cdot w_{2,1} = [1.0472, 3.1706]$ &&$[\overline{t}]_{2,1}$&= $[1, 1]$\\
       &$y$&= $[t]_{2,0} + [t]_{2,1} + b_{2,0} = [1.369, 3.703]\hspace{.2cm}$ &&$\overline{y}$&= $[1, 1]$\\
   \end{tabular}
    \end{adjustbox}

    \caption{Interval adjoint AD example on a NN with two inputs and one output. Two hidden layers with 2 nodes at each layer using $\text{ReLU}$ activation function. The NN is evaluated with interval input $[\vx]=([x]_0, [x]_1)^{\top}$, where $[x]_0 = [x]_1 = [1, 10]$. After the forward evaluation of the primals on the left, the adjoint operations on the right are evaluated in reverse.}\label{tab1}
\end{table}

    We aim to clarify the details of interval adjoints through the \autoref{ex-a}. The specific manner in which to use this information for pruning is discussed next.
 
\begin{example}\label{ex-a}	
		Consider $f(\vx) = \ln(x_0 \,\cdot\, x_1) + cos(x_0 \,/\, x_1)$ to be approximated by a small NN (2 hidden layers with $ReLU$ activation function on each layer). Let $[\vx] = ([1, 10], [1, 10])^{\top} \in [\R]^2$ be the input range. Interval arithmetic and AD is applied on this NN to compute interval adjoints. The computational graph  in the Figure ~\ref{fig_0} depicts a possible approach to evaluate the primal values of the NN through the forward pass of adjoint AD. The computational graph also illustrates the reverse pass process for computing the adjoints. Table ~\ref{tab1} shows  step by step evaluation of forward and reverse pass to evaluate interval primals and interval adjoints of two layer NN. The output of the NN is thus in the interval $[1.369,3.703]$. 
\end{example}

\subsection{Interval adjoint significance analysis (IASA)}\label{IASA}\label{sec:pruning-derivatives}
Deviations in outputs $\Delta{y}$ due to deviations in inputs $\Delta{x}$ are considered as sensitivity. It is essentially the difference between deviated and non-deviated outputs: $f((x+\Delta{x})*w)-f(x*w)$. In numerical computations, the sensitivity of the output with respect to the input parameters can be estimated using their gradient information. These sensitivities can also be measured for interval data using interval types.
	
	Research has been conducted on using network parameter sensitivities ~\cite{sensivity2,sensivity5} for pruning purposes. The sensitivity of input parameters is valuable information in defining the significance of each parameter in numerical computations as well as in neural networks. However, one issue with sensitivity-based methods in neural networks is that they rely entirely on trained data/samples to calculate gradient information. Additionally, it is costly to compute the gradient of parameters for each sample and then aggregate them. This problem can be solved by using intervals. Interval calculations always provide a guaranteed enclosure of the desired results.

	IASA ~\cite{IASAFF} is a method used to evaluate the significance of all the input and intermediate variables for numerical functions. It is based on the concept of interval types, which allows the measurement of sensitivities using gradient information for interval data. Adjoint AD is applied here which not only computes the primal values of numerical functions but it also computes the impact of individual input and intermediate variables on the output of an interval-valued function. The significance of an interval input variable $[x]$ can be defined as the product of the width $w([x]) = x_\top - x_\bot$ of the interval vector input $[x]$,  and the absolute maximum of the first-order derivatives of the interval output $[y]$ with respect to interval input $[x]$ ~\cite{riehme2015significance}.
	\begin{equation}\label{eq:significane-analysis-general} 
		S_{[y]}([x]) = w([x]) \cdot \max \bigl(\bigl|\nabla_{[x]} [y]\bigr|\bigr).
	\end{equation} 
 
    In \autoref{eq:significane-analysis-general},  $\nabla_{[x]} [y]$ is the interval adjoint of $[y]$ w.r.t. interval $[x]$ and the maximum absolute value of this interval adjoint is used.
	For a trained neural network, we consider the entire range of the training input data as the input interval and use it during significance analysis. However, in some applications, it might be useful to perform interval splitting ~\cite{deussen2016automation} due to unfeasible relational operators or the \textit{wrapping effect}. Trained weights and biases remain unchanged and are used to compute the interval outputs. This eliminates the need to compute derivatives for each trained sample and aggregates them, thereby making the process much more efficient. For a better understanding \autoref{eq:significane-analysis-general} is reformulated in the context of neural networks as:
	\begin{equation}\label{eq:significance-analysis-network} 
		S_{[y]}([n]_{l, i}) = w([n]_{l, i}) \cdot \max \bigl(\bigl|\nabla_{[n]_{l, i}} [y]\bigr|\bigr).
	\end{equation}
		
	If there is only one output of the NN, then \autoref{eq:significane-analysis-general} is used for IASA, where $l$ denotes the hidden layer number and $i$ denotes the node number in layer $n$. The width $w([n]_{0, i})$ measures the impact of interval input $[x]_i$ on first hidden layer's node $[n]_{0, i}$. The larger the width of node $[n]_{0, i}$, the larger the impact of the input $[x]_i$ and vice versa. But this information alone is insufficient in describing the significance of $[n]_{0, i}$ . Further operations (e.g. batch normalization) during the evaluation of the output and different intermediate nodes may increase or decrease the influence of node $[n]_{0, i}$. This problem is solved by computing the influence of the node $[n]_{0, i}$ on the overall output $[y]$ of the NN. Absolute maximum of the first order partial derivative $\max(|\nabla_{[n]_{0, i}}[y]|)$ of node $[n]_{0, i}$ measure the influence of node $[n]_{0, i}$ on the overall output $[y]$ of NN. If the absolute maximum value of derivative $\max(|\nabla_{[n]_{0, i}}[y]|)$ is small, a change in the value of node $[n]_{0, i}$ has small influence and vice verse. Therefore, the product of the width of a node and its absolute maximum first order derivative is a suitable criterion to define the significance of a node.
	
	IASA is applied to a small, already trained NN (as shown in Figure ~\ref{fig_0}), Example ~\ref{ex-a}	in section 4. \autoref{eq:significance-analysis-network} is used to determine the significance of each node in the neural network. Significance value of the first and second node at first hidden layer are $1.02$ and $1.31$ respectively. Both of the nodes are almost equally significant according to their significance values. The significance value of the first and second node at second hidden layer are $0.21$ and $2.12$ respectively. Node $[n]_{1,0}$ is less significant compared to node $[n]_{1,1}$ at the second hidden layer and can be safely removed from the NN. If node $[n]_{1,0}$  is removed, all the weights associated with it can also be removed. For clarity, the next insignificant node to be pruned, including its incoming at outgoing connections, are displayed in a shaded color in \autoref{fig_0}.

    Instead of just removing a node, IASA can use the midpoint of the outgoing connection from the removed node and add it to the bias of the next layer's node. In the above example, node $[n]_{1, 0}$, its incoming and out going connections up till next and previous layer are removed from the NN and the midpoint of the outgoing connection $[t]_{2, 0}$ is added to the bias of the next layer's node, which is the output node $[y]$. The mean squared error ($MSE$) of this pre-trained network was $0.0296531$ before applying IASA, and after removing a node from the NN using IASA, it decreased to $0.0272298$. In cases where IASA results in an increase in $MSE$ or a decrease in accuracy, the NNs can be fine-tuned with a small learning rate in the hope to recover the original accuracy/$MSE$.
\section{Case Study: Option Pricing of Gaussian Baskets}\label{sec:case-study}

As a case study, we consider the domain of options pricing and aim to find a neural-network surrogate model that approximates the true pricing function. In particular, we wish to find a surrogate model of the Bachelier model for pricing Gaussian basket options.
Although the problem itself can be effectively learned through a relatively small network by today's standards, it nonetheless highlights the core concepts of the presented methods and fundamental challenges one has to address for the methods to be applicable. A convenient property of this problem is that we can scale it up to arbitrary basket dimensions and can thus stress test the presented methods for varying basket sizes. 
Besides, the first-order (Delta) and second-order (Gamma) pricing sensitivities can be compared to analytical results when considering European payoff but are in general not analytically solvable for almost any other payoff function of interest. Recovering such sensitivities is of vital importance in the financial domain and a surrogate model that cannot provide the appropriate risk information is not acceptable.

We like to stress, however, that first and second-order sensitivity information is also important in many other areas, including engineering, physics, etc.
The methods are generally applicable, in particular when considering to predict conditional expectations of some stochastic process.

\subsection{Bachelier}

The Bachelier model can be described as a SDE:
\begin{equation}
\diff{S_t} = \mu S_t dt + \sigma\diff{W_t},
\end{equation}

where $t>0$, $\mu$ is the constant drift for the interest rate, $\sigma$ is the constant volatility, $S_t$ the underlying asset price at time t, and $\diff{W_t}$ describes a Wiener process, i.e. Brownian motion.

We can simplify this formula by considering the forward price.
\begin{definition}[Forward Price]
The forward price $F_t$ is the discounted price, i.e. \[F_t=S_t e^{r(T-t)},\]
where r is the interest rate and time T the maturity.
\end{definition}
Alternatively, when considering the T-forward measure $\mathcal{Q_T}$ \cite{glasserman2003}, the model simplifies to 
\begin{equation}
    \diff{F_t} = \sigma\diff{W_t}^{\mathcal{Q_T}}.
\end{equation}

The forward measure results in the drift term to disappear from the SDE as it is incorporated in the probability measure of the modified Wiener process.

\begin{definition}[European call option payoff]\label{def:europeanpayoff} The European call option payoff at maturity $S_T$ and strike $K$ is given by:
    \[\upnu(S_T, K) = (S_T - K)^+.\]
\end{definition}

\begin{definition}[Option price]
    The option price is the expected value of the payoff, i.e.:
    \[V = \E[\payoff(S_T, K)].\]
\end{definition}

Let $V_C(F_t, K)$ denote the price of the European call option at time t for strike K.
The call option price at time $t=0$ can be calculated analytically and is:

\begin{equation}
    \callprice(F_0, K) = (F_0 - K) \normalcdf(z) + \sigma \sqrt{T}\normalpdf(z),\quad z=\frac{F_0 - K}{\sigma \sqrt{T}},
\end{equation}

where $\normalpdf$, $\normalcdf$ are the PDF and CDF of the standard normal distribution, respectively. We will use the analytic prices for comparison with the prices obtained by the surrogate model. Although the price is known in the specific case of a European option, for more exotic payoffs, the analytic solution does not exist and thus numerical methods must be used instead. 
Note that the famous model by Black and Scholes is in fact very much related to the model of Bachelier, using a log-normal distribution instead of the normal distribution \cite{schachermayer2008bsvsbachelier}.
For further derivations of the Bachelier model, including its characteristic function and PDE description see, e.g., \cite{terakado2019option}.

\subsection{Gaussian Basket}

We will consider basket models with assets being jointly normal distributed. As a result, the price of the basket option will remain Gaussian and the Bachelier model computes suitable prices.
\begin{definition}[Basket]
    A Basket $\rmS_t \in \R^m$ of $m$ securities\\ $S^{[0]}_t, S^{[1]}_t, \ldots, S^{[m]}_t$ has at time $t$ the price:

    \[\rmS_t = \sum_{i=0}^{m} \omega_i S^{[i]}_t, \quad \sum_{i=0}^{m} \omega_i = 1,\]

    where $\omega_i$ is the weight associated with the $i$th security. 
\end{definition}
In particular, consider a basket with correlated normally distributed assets that can be generated from a multivariate normal distribution.
Then, we can model the basket option with a correlated Bachelier model for $m$ assets by

\begin{equation}
\diff\rmF_t\ = \bm{\sigma} \diff \rmW_t,
\end{equation}
where $\rmF_t \in \R^m$ and $\diff W_t^j \diff W_t^k = \rho_{jk}$ with $j,k \in \{1,\ldots, m\}$. That is, the Wiener process $\diff W_t^j$ is correlated to the process $\diff W_t^k$ with constant $\rho_{jk}$. For $j=k$, the correlation is $1$. Each asset $j$ has a volatility $\sigma_j$. So, the constant volatilities in $\bm{\sigma}$ are applied elementwise.

As pointed out by \citet{Huge2020}, although we deal with $m$ underlying assets, the basket option price will turn out to be a nonlinear function of a single dimension. It requires a surrogate model to perform large dimensionality reduction to uncover and correctly represent this pricing function.  

The Greeks can be found through differentiation via AD or analytically. As an example, we analytically derive the Delta of the European call option price:

\begin{align}
    \frac{\partial\callprice(F_0, K)}{\partial F_0} &= \frac{\partial}{\partial F_0}\Bigl[ (F_0 - K)\normalcdf(z)+\sigma\sqrt{T}\normalpdf(z) \Bigr]\nonumber\\
    &= \normalcdf(z) \frac{\partial}{\partial F_0} (F_0 - K) + (F_0 - K) \frac{\partial}{\partial F_0}\normalcdf(z) + \sigma\sqrt{T}\frac{\partial}{\partial F_0}\normalpdf(z)\nonumber\\
    &= \normalcdf(z) +  \frac{(F_0 - K)\normalpdf(z)}{\sqrt{2\pi}\sigma\sqrt{T}} -  \frac{1}{\sqrt{2\pi}}\normalpdf(z) \frac{F_0 - K}{\sigma \sqrt{T}}\nonumber\\
    &= \normalcdf(z), \quad z=\frac{F_0 - K}{\sigma \sqrt{T}}.\label{eq:bachelier:delta}
\end{align}

The Gamma can be analytically computed in a similar fashion. The analytical results are used as reference results, i.e. test data, for all upcoming experimental results. 

\section{Least-Squares Monte Carlo}\label{sec:least-squares-mc}

Computing an option price for each configuration of the initial parameters by sampling many payoffs and averaging over them through Monte Carlo sampling, is often too costly to perform. Instead of throwing away the sampled payoffs after finding the option price for some initial spot price $S_0$, could we instead reuse the prior generated information for predicting the price given other $S_0$ and hence save a large fraction of the computational cost? 

\citet{longstaff2001leastsquaresmc}, albeit in the context of American options, where amongst the first to reformulate the problem of pricing options given various input spot prices as a regression problem. We hereby sample from the possible input range uniformly at random and fit a curve between the output payoff samples through least-squares regression.
This Least-Squares Monte Carlo method can be formalized as optimizing

\begin{equation}\label{eq:leastsquares}
    \params^* = \argmin_{\params} \E_{(\vtheta, \vz) \sim \inparamsset \times \mathcal{Z}}  \Bigl[ {\lVert \payoff(f(\vtheta, \vz)) - \net(\vtheta) \rVert}^{2}_{2} \Bigr],
\end{equation}
where $\net$ is the fitted curve with coefficients $\params$ for random input parameters $\vtheta \sim \inparamsset$ and random path noise samples $\vz \sim \mathcal{Z}$.

This setup lends itself for use of neural networks as the regressor. 

\subsection{Regression using Neural Networks}\label{sec:regression-nn}

In both scenarios, we use a Multi-Layer Perceptron (MLP) as the surrogate model. The MLP has 6 hidden layers with width 128 and, uses the SiLU activation function, a continuously differentiable alternative to ReLU. The optimizer $G$ is chosen to be Adam with default settings as in \cite{KingmaB14}. In addition, we use a cosine one-cycle learning rate schedule with peak $\eta=0.1$, raising $\eta$ for $30\%$ of the cycle, starting at $\eta=\num{4e-3}$, and finishing with $\eta=\num{1e-5}$.

\begin{figure}[h]
    \centering
    \includegraphics[width=\linewidth]{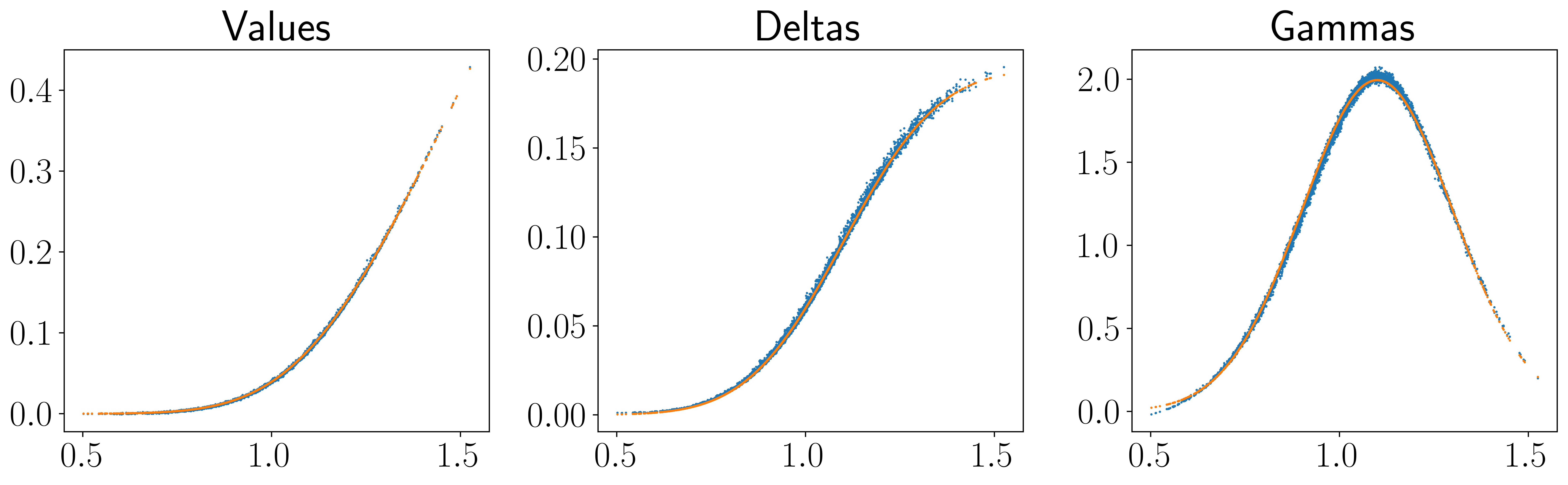}
    \caption{Baseline results of ML with a basic MLP.}
    \label{fig:baseline-mlp}
\end{figure}

The results in \autoref{fig:baseline-mlp} will serve as a baseline for all upcoming model pruning and surrogate modelling techniques. It consists of three plots, one for the price predictions (Values), the predictions of first-order sensitivities (Deltas), and the second-order sensitivities.

\subsection{Learning with pathwise derivatives}

Typically in ML, the problem of overfitting is addressed through some form of regularization. 
The Bayesian perspective highlights that we can view least-squares regression as Maximum-Likelihood estimation under the assumption of Gaussian noise being present in the output. Furthermore, $L_2$ regularization corresponds to Maximum-a-Posteriori estimation given furthermore a Gaussian prior on the weights \cite[Ch. 3.3]{bishop2006}.
However, such regularization adds additional hyperparameters leading to the well known bias-variance tradeoff.

We instead consider Gaussian noise in the input and realize that the loss has to depend on the gradient \cite{pmlr-v80-srinivas18a}. We can then apply Differential ML. In the next section, we highlight how to find such gradients in the context of Monte Carlo sampling. We thus are interested in the pathwise derivatives.

\subsubsection{Interchanging the derivative and expectation}

In order to take the derivative of the payoff function $\payoff$, we first consider the realization of the random path as an explicit parameter $\vz\sim \mathcal{Z}$, where the set $\mathcal{Z}$ represents the possible random vectors. The payoff can be decomposed into a function on $g$, where $g:\inparamsset\times\mathcal{Z}\rightarrow\outparamsset$ represents the underlying path:
\[
\payoff(g(\vtheta, \vz)),
\]
where the input parameters $\vtheta\in \inparamsset$. In the case of Bachelier, $\vtheta=(S_0, K)$, $f(\vtheta)=S_T - K$ and $\payoff$ is $(\cdot)^+$. E.g., $S_0\sim U(90, 110)$ and $K=100$. Alternatively, we can also think of $g$ returning multiple values that correspond to the parameters of the payoff function in \autoref{def:europeanpayoff}.

We have unbiased estimates of pathwise derivatives of the payoff, if:

\begin{equation}\label{eq:pathwise:unbiasedestimate}
\E_{\vz\sim \mathcal{Z}}\Bigl[\frac{\partial}{\partial S_0}\payoff(g(\vtheta, \vz))\Bigr] = \frac{\partial}{\partial S_0}\E_{\vz\sim \mathcal{Z}}\Bigl[\payoff(g(\vtheta, \vz))\Bigr].    
\end{equation}
So, if we can interchange the derivative with the expectation as above, it is possible to compute the derivative for each path individually. We will discuss the applicability of the method in Section \ref{sec:pathwise:applicability}.
The derivative of a path can further be broken down using the chain rule: 
\begin{equation}\label{eq:pathwise:derivbachelier}
\frac{\partial}{\partial S_0} \payoff(g(\vtheta, \vz)) = \frac{\partial \payoff(g(\vtheta, \vz))}{\partial S_T} \frac{\partial S_T}{\partial S_0}.
\end{equation}
We will be using adjoint AD to compute \autoref{eq:pathwise:derivbachelier} automatically.
As an example, we provide the analytic pathwise derivative for the Bachelier model of a European call option. 

\subsubsection{Bachelier}

We consider again the Bachelier model.
We get for a fixed $\vz\sim \mathcal{Z}$ and $\vtheta \in \inparamsset$:

\begin{equation}\label{eq:bachelierpayoffpathwisederivative}
    \frac{\partial\payoff(g(\vtheta, \vz))}{\partial F_T} = \frac{\partial}{\partial F_T} (F_T - K)^+ = \mathbb{1}_{F_T > K}.
\end{equation}
Note that at $F_T=K$ the derivative does not exist, but the event $F_T=K$ occurs with probability 0. As a result, the payoff function is almost surely differentiable with respect to $F_T$.

Furthermore, 

\[
F_T = F_0 + \int_{0}^{t} \sigma \diff{W_t}, \quad 0\leq t\leq T, \quad \text{(by definition of SDE)}
\]
So,
\[
\frac{\partial F_T}{\partial F_0} = 1.
\]
Overall the pathwise derivative of the payoff under the Bachelier model is just $\mathbb{1}_{S_T > K}$.
For the basket option, the pathwise derivative payoff for the individual dimensions remains the same since \[\frac{\partial F_t^{(i)}}{\partial F_0^{(j)}} = 0, \text{for all } i\neq j,\quad \frac{\partial F_t^{(i)}}{\partial F_0^{(i)}} = 1.\]
However, the pathwise derivative is not always well-defined.

\subsubsection{Applicability}\label{sec:pathwise:applicability}

The method of pathwise derivatives is only applicable if certain conditions can be fulfilled. \citet[393--395]{glasserman2003} discusses practical sufficient conditions to verify the validity of the pathwise method.
In practice, the deciding criterion is whether the (discontinuous) payoff function $\payoff$ is Lipschitz continuous with respect to the parameters, and differentiable almost everywhere.

\begin{definition}\label{def:pathwisederivlipschitz}
The payoff function $\payoff$ is Lipschitz continuous, if there exists a real constant $k_{\payoff}\geq 0$ such that for all $\vtheta_1, \vtheta_2 \in \inparamsset$ and $\vz \sim \mathcal{Z}$

\begin{equation}
\|\payoff(g(\vtheta_2, \vz)) - \payoff(g(\vtheta_1, \vz))\| \leq k_{\payoff} \| \vtheta_2 - \vtheta_1 \|,    
\end{equation}
i.e. it adheres to the Lipschitz continuity condition. 
\end{definition}

If $\payoff$ is a smooth function of $f$ it is sufficient to consider whether $f$ is Lipschitz continuous. However, we almost always deal with non-smooth payoff functions like in the next example.

\begin{example}
    Consider the European call option payoff. We already discussed at \autoref{eq:bachelierpayoffpathwisederivative} that the payoff has only one non-differentiable point at $S_T=K$ and is thus almost everywhere differentiable. Furthermore, the payoff is Lipschitz continuous because the $(\cdot)^{+}$ function is Lipschitz continuous:

    Let $y_1, y_2 \in \outparamsset$ represent the output of two different paths and assume w.l.o.g. that $y_1 \geq y_2$,

    \begin{align}
            \| \payoff(y_1) - \payoff(y_2) \| &= \| (y_1)^+ - (y_2)^+\| \\
            &= \left\{\begin{array}{lr}
        \|y_1-y_2\|, & \text{if } y_1>0, y_2>0\\
        \|y_1\|, & \text{if } y_1 > 0, y_2 < 0\\
        0, & \text{if } y_1 < 0, y_2 < 0
        \end{array}\right\}\\
        &\leq \| y_1 - y_2 \|.
    \end{align}
\end{example}

If we are, in addition, interested in second-order pathwise derivative information, we further require $\payoff$ to be twice differentiable almost everywhere and that \autoref{def:pathwisederivlipschitz} holds for $\payoff^{(1)}$. To be precise, there exists a real constant $k_{\payoff^{(1)}}$ such that for all $\vtheta_1, \vtheta_2 \in \inparamsset$ and $z \sim \mathcal{Z}$

\begin{equation}\label{eq:lipschitz}
\|\payoff^{(1)}(g(\vtheta_2, \vz)) - \payoff^{(1)}(g(\vtheta_1, \vz))\|  \leq k_{\payoff^{(1)}} \| \vtheta_2 - \vtheta_1 \|.    
\end{equation}
These conditions can be generalized to higher-order pathwise derivatives by requiring $\payoff$ to be $n$ times differentiable almost everywhere and taking the $(n-1)$th derivative of $g$ in \autoref{eq:lipschitz}.

\begin{example}
    Again, the European call option payoff, but now considering applicability for second-order pathwise derivatives. We know that $\payoff^{(1)}$ is a Heaviside step function which is clearly not Lipschitz continuous as it is not even continuous. Therefore, \autoref{eq:pathwise:unbiasedestimate} does not hold. Without modification, second-order pathwise derivatives are thus not applicable. This turns out to be almost always the case in options pricing since the European payoff is amongst the simplest payoffs to be considered.
    Note that without the Lipschitz condition, we have that $\payoff^{(1)}$ is differentiable everywhere except at the strike and thus almost everywhere differentiable. However, the value of the derivative is always 0 whenever it does exist. Almost everywhere differentiable payoff functions are thus not sufficient.
    The nature of the Dirac delta is not captured when considering the pathwise method. The same problem occurs in more exotic options, e.g. Barrier payoffs.
\end{example}

The last example is motivation to consider techniques to make the second-order pathwise derivatives applicable by modifying the payoff functions to be well-behaved. A technique that is often used in this context is smoothing.

\subsubsection{Smoothing payoff and activation functions}
To alleviate the problem of discontinuous payoff functions we consider smoothing the payoff function. Many smoothing functions have been proposed but we only consider sigmoidal smoothing. 
The activation function of the MLP which is often chosen to be ReLU can also benefit from smoothing for better implicit predictive power of the Deltas ($\frac{\partial V}{\partial S_0}$). Applying sigmoidal smoothing to ReLU leads to the SiLU activation function.

\begin{figure}[h]
\begin{center}
    \begin{tikzpicture}[scale=0.75] 
      \begin{axis}[
        xlabel=$x$,
        ylabel=$f(x)$,
        xmin=-.5, xmax=1.5,
        ymin=-0.1, ymax=1.1,
        grid=both,
        samples=1000,
        domain=-.5:1.5,
        legend style={at={(0.7,0.2)},anchor=north}
      ]
    
      \def\p{0} 
      \def\w{0.05} 
    
      \addplot[myorange, thick, dashed] {        
        x < 0 ? 0 : x
      };
      \addplot[myblue!60!black, thick] {x/(1+exp(-((x-\p)/\w)))};
      \addlegendentry{Reference function}
      \addlegendentry{Sigmoidal smoothing}
      \end{axis}
    \end{tikzpicture}
\end{center}
\caption{Smoothing functions for payoff $(\cdot)^{+}$, where smoothing width $w=0.05$.}
\label{fig:smoothing}
\end{figure}
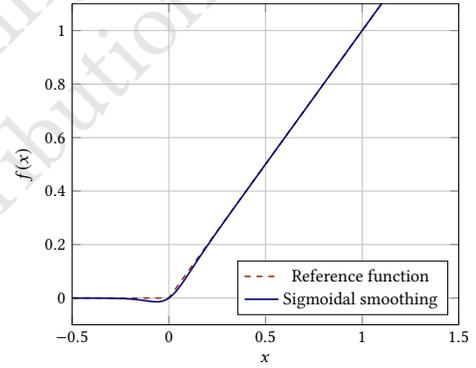

\paragraph{Sigmoidal Smoothing}
A general approach for smoothing the discontinuous transition between two functions at position $p$ can be achieved through the use of sigmoidal smoothing.
We perform smoothing between function $f_1: \R \rightarrow \R$ and $f_2: \R \rightarrow \R$ via $\Tilde{f}: \R \times \R^2 \rightarrow \R$ defined as
\[
\Tilde{f}(x, p, w) = (1 - \sigma(x, p, w))f_1(x) + \sigma(x, p, w)f_2(x),
\]
where 
\[
\sigma(x,p,w)=\frac{1}{1+e^{-\frac{x-p}{w}}},
\]
and $p$ is the position to change between the two functions and $w$ the width of the smoothing.
For $\payoff=(\cdot)^+$, we can first split up the function into $\begin{cases}
    0, & x < 0\\
    x, & x \geq 0
  \end{cases}$. By sigmoidal smoothing we thus get:
\[
\Tilde{\payoff}(x, w) = \frac{x}{1+e^{-\frac{x}{w}}}.
\]

\section{Results}\label{sec:results}
Throughout this paper, we use a consistent color scheme for visualization. The color \textcolor{truecolor}{orange} is used to represent true values, while \textcolor{predictedcolor}{blue} denotes predicted values.

\subsection{Results of model pruning using IASA}

In the initial phase of our experimentation, a regression model based on the Bachelier model is trained with six hidden layers, each consisting of 128 nodes, and applying the SiLU activation function. This model served as our baseline (see \autoref{fig:baseline-mlp}), and subsequently, IASA is performed on each of the layers to assess the significance of the nodes. The goal is  to identify and prune less significant nodes to optimize the network architecture. 

\begin{figure}[h]
    \centering
    \includegraphics[width=\linewidth]{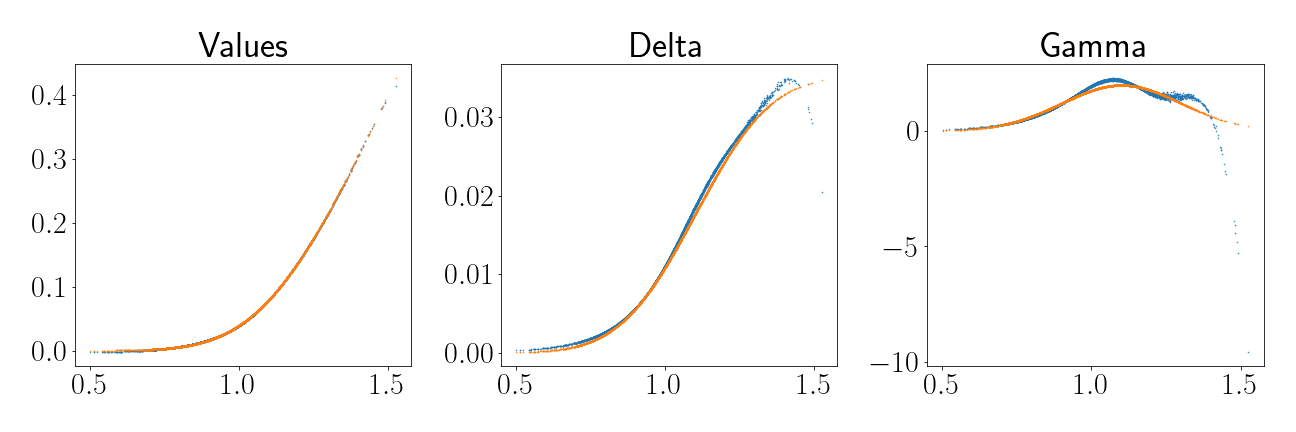}
    \caption{Results of the pruned model.}
    \label{fig:pruned-NN}
\end{figure}

The NN is pruned iteratively, layer wise, and and after the each round of pruning the NN is retrained. In the initial pruning cycles only 10 epochs and a small training data set is enough to recover the value loss. Notably, IASA demonstrated its efficacy by removing a substantial number of nodes while maintaining the model's predictive pricing accuracy.
Specifically, after the pruning process, the network attained similar $R^2$ score (i.e., the coefficient of determination) with only three nodes at both the first and second dense layers. The number of nodes at the third layer was reduced to one, and subsequent layers retained only two nodes. 
The resulting surrogate model achieves comparable value accuracy with significantly fewer nodes. The result can be seen in Figure ~\ref{fig:pruned-NN}.

\begin{figure}[h]
    \centering
    \includegraphics[width=\linewidth]{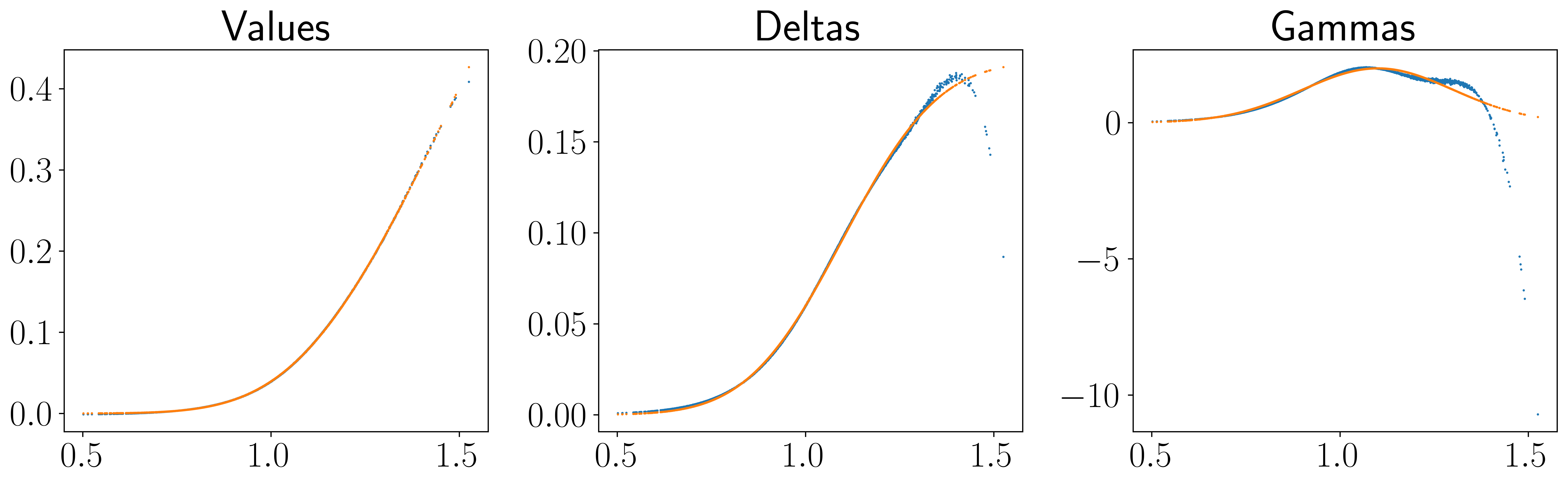}
    \caption{Results of pruned model after pruning the layer 3, 4 and 5 and retraining.}
    \label{fig:pruned-NN-layers}
\end{figure}

Further reduction in model complexity is explored by evaluating the impact of removing entire layers from the pruned architecture. It is observed that retaining only a single node at the third layer allowed for the removal of subsequent hidden layers without compromising pricing performance. The network is retrained again using the remaining weights after pruning the layer three, four and five, recovering any potential loss in predictive accuracy (see \autoref{fig:pruned-NN-layers}). 

\begin{figure}[h]
    \centering
    \includegraphics[width=\linewidth]{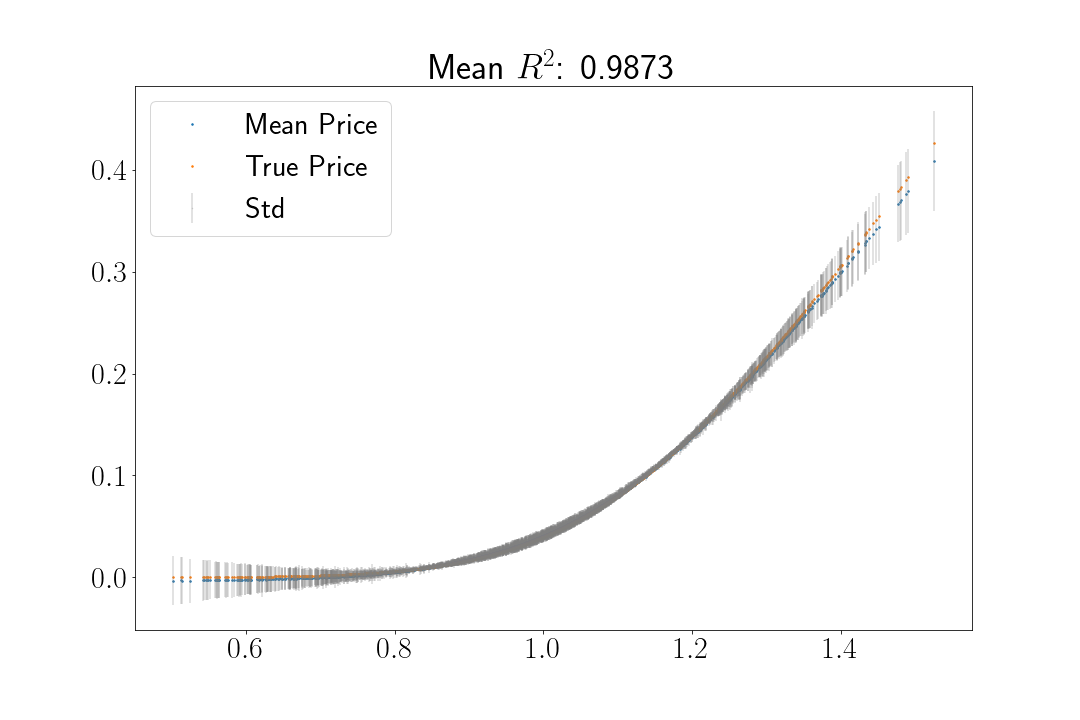}
    \caption{Pruned architecture trained with random weights.}
    \label{fig:random_trained}
\end{figure}

To illustrate the effectiveness of the pruning process, a comparison is made with a network architecture randomly initialized with three nodes at each of the two dense layers. Notably, achieving a network with only three nodes at two layers through random initialization proved to be an exhaustive and challenging task. The results highlighted in the plot in the Figure ~\ref{fig:random_trained} showcase the true mean and standard deviation, providing insight into the distribution of prediction outputs. In contrast, the pruning approach using interval adjoints can systematically reduce the network's complexity from having 128 nodes at each layer to an optimized configuration with minimal nodes. We can thus use IASA to find appropriately sized surrogate models. However, the accuracy in recovering the sensitivities, here Delta and Gamma, is fading. It is a shortcoming of all currently known pruning methods as they solely focus on the value prediction accuracy. We next show that Sobolev fine-tuning can overcome those shortcomings.

\subsection{Results after Sobolev fine-tuning}
Previously, practitioners mostly experimentally guessed an appropriate surrogate model size for learning with Differential ML.
From the interval adjoint based pruning we get a solid foundation for answering how large the surrogate model should be. The pruned model is then further fine-tuned in a final stage with Sobolev Training, as described in \autoref{sec:diffml}.

\begin{figure}[h]
    \centering
    \includegraphics[width=\linewidth]{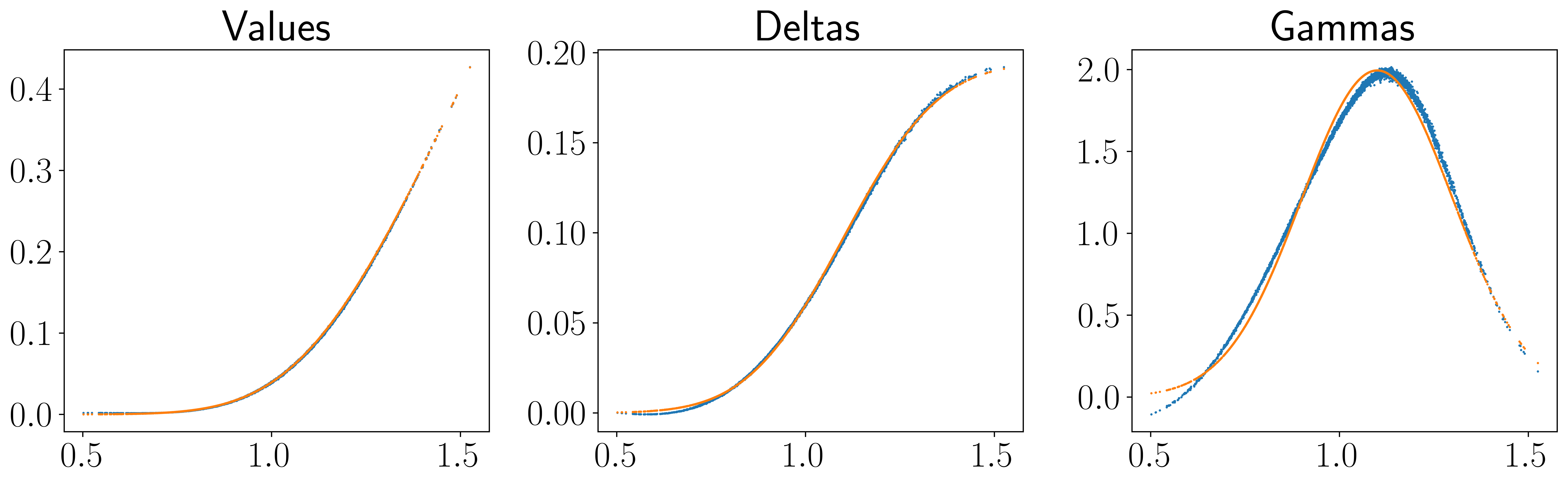}
    \caption{Results after Sobolev fine-tuning on derivative samples from learned neural network.}
    \label{fig:diffml-nn-mlp}
\end{figure}

After training a pruned model again with differential data from the large neural network, the surrogate model can recover the lost representation for the sensitivities. Note that we thus do not require access to the originating reference model. Delta predictions went from 0.998700 to 0.996718 during pruning with severe errors at the boundaries of the input domain. Now the surrogate model even surpasses the larger neural network in its accuracy of the Delta ($R^2=0.999479$), as it can learn from derivatives, highlighting that the fundamental shape of the curve is again represented in the surrogate. Similarly, the Gamma values previously dropped from 0.997033 to 0.902470 and after Sobolev fine-tuning has again an accuracy of $R^2=0.987393$. If we furthermore have access to the original model, here the Bachelier basket model, the accuracy of the predicted derivatives improves further (Delta: 0.999863, Gamma: 0.997374). It highlights that differential data from a learned larger network will not be as accurate as the ground truth model.
\autoref{fig:teaser} captures all results at once for direct visual comparison.

\begin{figure}[h]
    \centering
    \includegraphics[width=\linewidth]{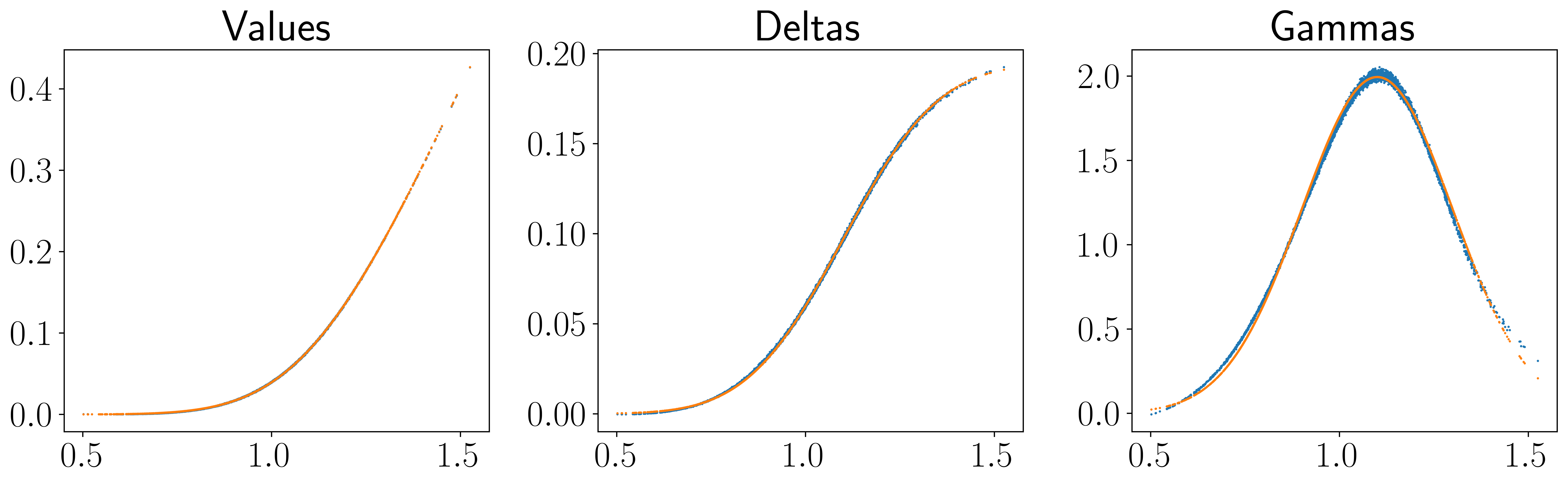}
    \caption{Results after Sobolev fine-tuning on derivative samples from Bachelier reference model.}
    \label{fig:diffml-bachelier-mlp}
\end{figure}

However, when performing Sobolev fine-tuning for the pruned network where layers have been removed, the resulting predictions will again be worse, following a very similar curve as \autoref{fig:diffml-nn-mlp}. It highlights that a too small network will eventually degrade in performance, even after being retrained with differential data.
Finally, the $R^2$ score of all methods is summarized in \autoref{fig:table:numbers}.

\begin{table}[h]
    \centering
    \renewcommand{\arraystretch}{1.5}
    \caption{$R^2$ score of surrogate models for predicting the price, deltas, and gammas of a Bachelier modelled basket option.}
    \begin{tabular}{@{}rrrrr@{}}
    \toprule
    \multirow{2}{*}{Predict} & \multirow{2}{*}{Oversized} & \multirow{2}{*}{Pruned} &  \multicolumn{2}{c}{Sobolev fine-tuning} \\
    \cmidrule{4-5} & \multicolumn{1}{c}{NN} & \multicolumn{1}{c}{NN} & \multicolumn{1}{c}{NN Data} & \multicolumn{1}{c}{Bachelier} \\
    \hline
    Values & 0.999545 & 0.999296 & 0.999805 & \textbf{0.999962} \\
    Deltas & 0.998700 & 0.996718 & 0.999479 & \textbf{0.999863} \\
    Gammas & 0.997033 & 0.902470 & 0.987393 & \textbf{0.997374} \\
    \bottomrule
    \end{tabular}
    \label{fig:table:numbers}
\end{table}

\section{Conclusion and Outlook}
In this paper, we improve upon existing surrogate modelling techniques by incorporating sensitivity information throughout the entire surrogate learning procedure. Starting from a larger, potentially already existing neural network, interval adjoint significance analysis efficiently removes neurons and thus prunes the network down to its critical size. 
The implicitly encoded sensitivity information, i.e. the (second-order) derivatives, is accurately modelled after fine-tuning the surrogate model with differential data. We hereby use Sobolev Training, recovering the sensitivity information after only a few fine-tuning epochs.
We end up with a justifiably small and derivative-informed surrogate model for use in efficient pricing and risk assessment of Basket options. However, the proposed method is applicable beyond option pricing. The highlighted methods directly extend to any domain where a conditional expectation must be found from a stochastic process, modelled through a stochastic differential equation. Challenges in the application of pathwise derivatives can be overcome through smoothing and thus serves as a general-purpose efficient differential sampling method. Furthermore, the application of algorithmic differentiation is far-reaching and allows for sampling derivative information of arbitrary reference models or neural networks. The proposed method should thus serve as a general recipe for finding appropriately sized surrogate models that recover sensitivity information.

We do not consider pruning edges of the surrogate network to not result in a sparse model, as current hardware is still ill-equipped for executing sparse operations. Nonetheless, extensions to  
significance analysis on the edge level could provide meaningful further insights. We also observed that in this particular example, second-order differential information was not needed for recovering the Gamma predictions. The effect of second-order differential data specifically for recovering second-order sensitivities of a pruned neural network surrogate model on other problem domains requires further investigation.

\bibliographystyle{ACM-Reference-Format}
\bibliography{references}

\appendix

\end{document}